\documentclass[letterpaper,journal]{IEEEtran}
\ifdefined\pdfminorversion\pdfminorversion=7\fi
\usepackage{amsmath,amssymb,bm}
\usepackage{graphicx,booktabs,array,cite,url}
\graphicspath{{figures/}}

\newcommand{\ind}{\mathbf{1}}
\newcounter{algorithm}
\begin{document}
\thispagestyle{empty}
\title{DynaRC: Dynamics-Robust 4-D Radar Relocalization\\
with Rate-Adaptive Sequence Matching}
% \author{anonymous
% \thanks{anonymous}}%
\author{Chuan Cao, Xingyi Li, Tianchen Deng, and Jingchuan Wang*,~\IEEEmembership{Senior Member,~IEEE}%
\thanks{Chuan Cao, Xingyi Li, Tianchen Deng, and Jingchuan Wang are with the School of Automation and Intelligent Sensing, Institute of Medical Robotics, Shanghai Jiao Tong University, Shanghai 200240, China, and the Key Laboratory of System Control and Information Processing, Ministry of Education of China, Shanghai 200240, China. *Corresponding author: Jingchuan Wang (e-mail: jchwang@sjtu.edu.cn).}}
\maketitle
\pagestyle{empty}       % 关闭后续所有页面的页码
\thispagestyle{empty}

\begin{abstract}
Four-dimensional millimeter-wave radar measures radial velocity within a
single scan and remains resilient under adverse weather and visibility.
Yet sparse directional returns, moving objects and occlusion make place
matching ambiguous. We propose DynaRC, a dynamics-robust framework encompassing
single-frame retrieval and sequence relocalization. Doppler evidence separates static returns and masks
motion-affected cells in a polar descriptor; median ring keys retrieve
candidates. Its sequence variant, Seq\_DynaRC, couples endpoint-anchored rate alignment and candidate-index voting
with ordered verification to associate the latest scan under unequal
map/query advancement, without odometry or cloud accumulation.
Across four indoor and outdoor map--query pairs, Seq\_DynaRC achieves
the highest conditional localization success within 2\,m among tested methods.
On a campus-road revisit under changing traffic, success reaches 91.8\%, exceeding
descriptor-matched SeqSLAM by 55.0 percentage points when each method retains
all its published outputs; output availability is reported separately.
Code and indoor/outdoor dynamic 4D radar data are available.
\end{abstract}


\section{Introduction}
Global relocalization recovers a robot's pose in a known map without a
reliable initial estimate. It comprises \emph{place recognition}, which
outputs candidate map locations, and \emph{pose recovery}, which estimates
the robot's six-degree-of-freedom (6-DoF) pose. Our contributions focus on
place recognition. Four-dimensional radar measures range, azimuth,
elevation and radial velocity, supporting localization under adverse
visibility~\cite{NTU,SPR}. Sparse returns, dynamics, occlusion and repeated
structures nevertheless create false matches; a low descriptor distance
alone cannot establish a correct global pose.

Two challenges arise. Within a scan, limited field of view and occlusion
remove static structure, so empty descriptor cells need not represent
observed free space. Across a sequence, vehicle speed, keyframe density and
query sampling alter map/query index correspondence. Fixed one-to-one
progression can associate the latest scan with a map frame ahead of or
behind its location. Temporal matching must therefore preserve both
consistency and the identity of the observation being localized.

DynaRC denotes the overall framework, comprising single-frame retrieval and
its sequence extension, Seq\_DynaRC. Their shared descriptor encodes a polar
height or intensity/RCS channel according to the environment and masks
cells with strong Doppler-derived dynamic evidence.
Median ring keys retrieve candidates for masked descriptor comparison.
Seq\_DynaRC introduces \emph{endpoint-anchored rate alignment} (ERA): each
hypothesis specifies the latest query's map endpoint and the window's index
advancement. Candidate-index voting selects supported hypotheses for aligned
multi-frame refinement. Iterative closest point (ICP) then registers the
current cloud to the chosen keyframe. Association requires neither odometry
nor inter-frame cloud accumulation.

The contributions are:
\begin{enumerate}
\item A training-free radar representation combining static polar features,
Doppler-based rejection of motion-affected cells, and median aggregation
to reduce extreme-cell influence during retrieval.
\item A candidate-supported sequence search coupling endpoint/rate proposals,
index-tolerant voting, and ordered verification to associate the current
scan under unequal map/query advancement.
\item An evaluation on four map--query pairs, including a descriptor-matched
SeqSLAM baseline, that reports localization success, output availability,
sampling differences, and computational cost.
\end{enumerate}

\begin{figure*}[t]
\centering
\includegraphics[width=\textwidth]{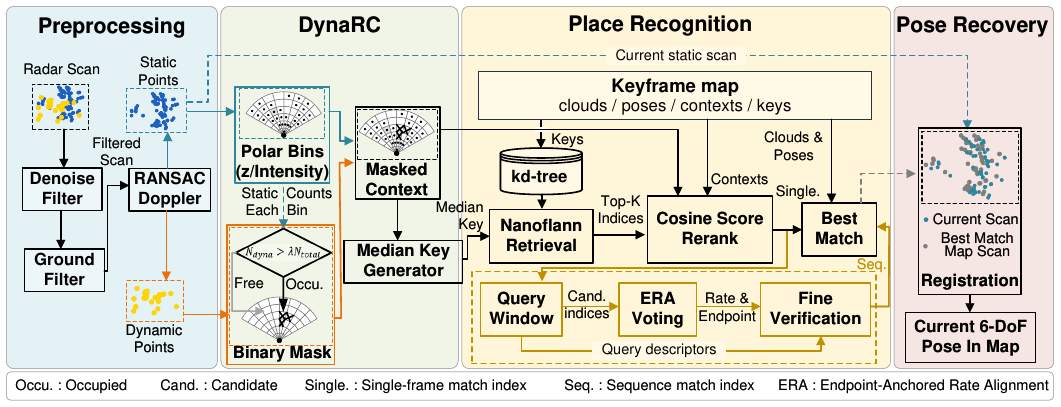}
\caption{DynaRC framework for radar relocalization.
Denoising and ground filtering precede Doppler-based separation of
static and dynamic returns. Static returns form polar feature contexts,
while per-cell static and dynamic counts determine a binary mask.
Median ring keys support kd-tree retrieval, followed by masked
cosine-distance reranking. The single-frame branch selects a matched
keyframe index; the sequence branch combines endpoint-anchored rate
alignment (ERA) voting with fine verification to select the map endpoint
associated with the current scan. Pose recovery registers the current
static scan to the selected keyframe and uses its map pose to recover
the current 6-DoF pose.}
\label{fig:framework}
\end{figure*}

\section{Related Work}
Relocalization methods differ in sensing coverage, scene representation,
and use of temporal information~\cite{PRSurvey}.

\subsection{Single Scan Radar Relocalization}
For omnidirectional scans, Scan Context and its extensions encode polar or
Cartesian structure for retrieval and viewpoint alignment~\cite{SC,SCpp};
MulRan adapts intensity-based Scan Context to radar~\cite{MulRan}. RaPlace
and Open-RadVLAD use signal transforms and aggregation~\cite{RaPlace,RadVLAD},
while ReFeree uses compact feature/free-space statistics~\cite{ReFeree}.
Robustness to noise or complete-scan transformations, however, does not
ensure robustness to structure lost through limited field of view (FOV)
or occlusion.

Learned automotive radar methods combine geometry and radar cross section
(RCS) through dynamic-point filtering, pointwise weighting or local--global
modeling~\cite{AutoPlace,SPR,TransLoc4D,RadarPR}. Removing moving points does
not recover static structure hidden behind them. mmPlace retains richer
signals through intermediate-frequency heatmaps and expands FOV using a
rotating platform~\cite{mmPlace}, unlike a single directional scan.

Visual and LiDAR methods use learned global aggregation or bird's-eye-view
features~\cite{NetVLAD,PointNetVLAD,BEVPlace}; SOLiD and UniLGL additionally
address restricted FOV~\cite{SOLiD,UniLGL}. Transfer to radar must account for
sparsity and RCS variability. Multimodal methods add complementary evidence
at the cost of additional sensing or training~\cite{PRFusion,LRFusionPR}.

\subsection{Sequence Relocalization}
Temporal methods increase geometric overlap or matching consistency.
Density-map loop detection accumulates local clouds~\cite{DensityMaps};
SPOT combines stereo-odometry accumulation with sequence matching for
similar and opposing viewpoints~\cite{SPOT}. Accumulation requires locally
consistent motion estimates. SeqSLAM instead searches coherent paths in
a locally contrast-enhanced query--reference difference matrix, allowing
different traversal rates~\cite{SeqSLAM}. Sequence consistency reduces
perceptual aliasing, while the path model defines which correspondences
are admissible. Reverse traversal requires explicit treatment, as in SPOT.

iLoc~\cite{iLoc} combines learned, attention-enhanced visual descriptors
with coarse-to-fine global feature matching. Candidate-level association
is followed by local refinement, while visual odometry and loop closures
support pose-graph state estimation. This separates efficient global search
from accurate pose recovery, but its visual features and motion-estimation
pipeline differ from radar-only descriptor-sequence association. UniPR-3D~\cite{deng2026unipr} aggregates 2D appearance and
geometry-grounded 3D tokens from a visual geometry transformer
for single-frame and variable-length sequence retrieval.
This learned multi-view representation complements approaches
that establish temporal consistency through explicit frame
correspondences.

For 4D radar, Hilger \emph{et al.} combine Doppler--IMU submaps, appearance
and odometry constraints, sequence filtering and learned introspective
verification~\cite{Hilger}. Their treatment of opposing views and false
closures illustrates the value of temporal support and verification,
with additional motion-estimation and training dependencies.

Seq\_DynaRC uses motion-aware radar evidence to retrieve candidate places
without geometrically aligned scan accumulation. Candidate-derived
endpoint/rate hypotheses restrict the search, index-tolerant voting rejects
unsupported windows, and ordered verification preserves the association
with the current scan. These stages combine efficient proposal generation
with temporal consistency before final registration.

\section{Problem Formulation and Single-Frame DynaRC}
\label{sec:single}
Let the map be
\begin{equation}
\mathcal M=\{(P_j,T_j,H_j,M_j,\mathbf r_j)\}_{j=0}^{N-1},
\end{equation}
where $P_j$ is a radar cloud in its keyframe coordinate system,
$T_j=T_{\mathrm{map}\leftarrow P_j}\in\mathrm{SE}(3)$ its global pose,
$H_j$ the feature context, $M_j$ its mask, and $\mathbf r_j$ the ring key.
Indices preserve acquisition order after invalid-frame removal; original
IDs are retained for reporting. $T_{A\leftarrow B}$ maps homogeneous
coordinates from $B$ to $A$. Given scan $Q_t$ and optional window
$\mathcal W_t=(Q_{t-L+1},\ldots,Q_t)$, retrieval identifies a map neighborhood
and registration estimates the residual rigid transform to recover the
query pose. Success is evaluated in XY; registration uses 3D clouds.

\subsection{Polar Feature Encoding and Dynamic Mask}
The map represents static structure. After denoising, Doppler random sample
consensus (RANSAC) estimates radar ego-velocity from measured radial
velocities and line-of-sight projections of sensor motion~\cite{RaVISLAM}.
Returns with Doppler residuals below a threshold form the static consensus;
inconsistent returns provide dynamic evidence (Fig.~\ref{fig:framework}).
Static points supply features; both sets supply
mask counts. Channel $c$ is either shifted height $f_z(p)=z(p)+h_0$,
$h_0$ being the radar mounting height above the ground, or recorded
intensity $f_I(p)$ (RCS may also be used).
For static returns $\mathcal B_{uv}$ in ring $u$, sector $v$,
\begin{equation}
H(u,v)=
\begin{cases}
\displaystyle\max_{p\in\mathcal B_{uv}} f_c(p), & \mathcal B_{uv}\ne\varnothing,\\
0, & \mathcal B_{uv}=\varnothing .
\end{cases}
\label{eq:height}
\end{equation}
$H(u,v)$ represents the strongest feature value within the cell.
The channel is fixed within a comparison; $z<-h_0$ retains negative shifted
height. Polar encoding organizes geometry or reflectivity, while motion-based
masking and robust aggregation provide dynamic robustness.

Let $n_s(u,v)$ and $n_d(u,v)$ count static and dynamic returns in each
cell, and let $n_{\mathrm{total}}=n_s+n_d$ be the total point count.
The binary mask is
\begin{equation}
M(u,v)=\begin{cases}
0,&n_s>0,\ n_d>n_{\min},\ n_d>\lambda n_{\mathrm{total}},\\
1,&\text{otherwise}.
\end{cases}
\label{eq:mask}
\end{equation}
Here $n_{\min}$ is the minimum dynamic-point count used to suppress
isolated-noise effects, and $\lambda$ is the within-cell dynamic-fraction
threshold. Empty static bins retain $H=0,M=1$: eligibility does not imply
observed free space. The mask rejects motion-affected cells without
propagating along rays or reconstructing occluded surfaces.

\subsection{Robust Ring Keys for Candidate Retrieval}
For ring $u$, collect the nonzero, unmasked observations, retaining
repeated values:
\begin{equation}
\mathcal Z_u=(H(u,v))_{v:\,M(u,v)=1,\ H(u,v)\ne0}.
\end{equation}
The compact key is
\begin{equation}
r_u=
\begin{cases}
\operatorname{median}(\mathcal Z_u),&|\mathcal Z_u|>0,\\
0,&|\mathcal Z_u|=0.
\end{cases}
\label{eq:ring}
\end{equation}
For even counts, the median averages the two central values. For nonzero
cell value $X$ and retention indicator $B$, value-independent thinning gives
$\Pr(B=1\mid X)=p>0$, hence $\Pr(X\le x\mid B=1)=\Pr(X\le x)$.
This preserves the population median, not necessarily an unchanged or
unbiased sample median. Its asymptotic 50\% replacement breakdown
point~\cite{RobustMedian} concerns extreme contamination, not invariance
below 50\% occlusion. The median is a robust default; structured occlusion
can violate the thinning assumption.

Ring keys are indexed with a kd-tree using Euclidean distance,
\begin{equation}
d_R(q,j)=\|\mathbf r_q-\mathbf r_j\|_2 .
\label{eq:ringdist}
\end{equation}
Joint sector permutation leaves the key unchanged, but changes in visible
surfaces can still alter it. Detailed comparison retains sector structure.
Single-frame retrieval uses five candidates; the sequence branch retrieves
up to 25 before reranking.

\subsection{Masked Descriptor Comparison}
The radar covers $120^\circ$, so column shifts use blank padding rather
than wraparound. Let $H_j^\ell$ and $M_j^\ell$ be the descriptor and mask
shifted by $\ell\in\{-(N_s-1),\ldots,N_s-1\}$ columns; out-of-view columns
have $H_j^\ell=M_j^\ell=0$ and do not contribute.
For query $q$ and sector $v$, define the jointly eligible rings
\begin{equation}
\mathcal I_v^\ell=\{u:M_q(u,v)=M_j^\ell(u,v)=1\}.
\label{eq:eligible}
\end{equation}
The vectors $\mathbf a_v^\ell=(H_q(u,v))_{u\in\mathcal I_v^\ell}$
and $\mathbf b_v^\ell=(H_j^\ell(u,v))_{u\in\mathcal I_v^\ell}$
use these rings in the same order. Unlike ring-key construction, this
comparison retains zero entries. A sector is usable when
$|\mathcal I_v^\ell|\ge2$ and both vector norms exceed a numerical-stability tolerance $\epsilon$.
Let $\mathcal V_\ell$ collect the usable sectors. The distance is
\begin{equation}
d_\ell(q,j)=1-\frac{1}{|\mathcal V_\ell|}
\sum_{v\in\mathcal V_\ell}
\frac{(\mathbf a_v^\ell)^\top\mathbf b_v^\ell}
{\|\mathbf a_v^\ell\|_2\|\mathbf b_v^\ell\|_2}.
\label{eq:masked}
\end{equation}
If no sector is usable, the implementation returns distance one.
Sector keys, obtained by averaging cell values over rings, give an initial
shift by minimum Euclidean distance. A local set $\mathcal L$ around
that shift is then searched:
\begin{equation}
d(q,j)=\min_{\ell\in\mathcal L}d_\ell(q,j).
\label{eq:distance}
\end{equation}
We retain $d(q,j)$ and its minimizing shift $a(q,j)$, choosing the smallest
shift on ties. Usable sectors contribute equally. Single-frame
retrieval selects the minimum-distance candidate, subject to a threshold
before registration.

\section{Seq\_DynaRC: Rate-Adaptive Sequence Relocalization}
\label{sec:sequence}
\begin{figure}[t]
\centering
\includegraphics[width=\columnwidth]{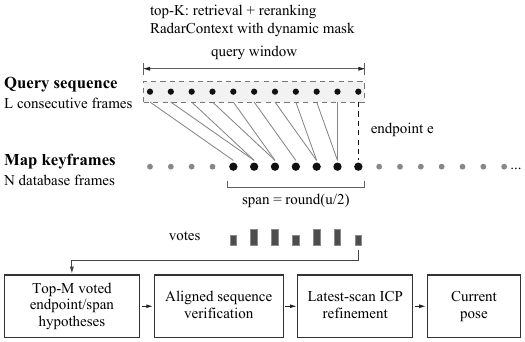}
\caption{Seq\_DynaRC. Each of $L$ queries retrieves top-$K$ candidates.
Solid links show repeated map indices; the dashed link anchors the newest
query to $e$. The span is $\operatorname{round}(u/2)$ index steps ($u$:
half-frame units). Votes select top-$M$ endpoint/span hypotheses for aligned
verification and latest-scan ICP. Frame counts and vote heights are schematic.}
\label{fig:sequence}
\end{figure}

Inspired by iLoc's coarse-to-fine global feature matching~\cite{iLoc},
we design a radar sequence matcher combining candidate-index voting
and aligned descriptor verification. ERA accommodates unequal map/query
advancement while anchoring the newest observation to the hypothesized
map endpoint (Fig.~\ref{fig:sequence}). Algorithm~\ref{alg:seq_dynarc}
summarizes sequence matching and current-pose recovery.

\subsection{Endpoint-Anchored Rate Hypotheses}
Reindex the chronological window as $Q_0,\ldots,Q_{L-1}$, with
$Q_{L-1}=Q_t$, and the searchable map as $P_0,\ldots,P_{N'-1}$, where
$N'=N-E$ excludes a configured database suffix. Ring-key retrieval and
masked reranking yield up to $K$ candidates $\mathcal C_i$ per query,
retained for temporal association even if single-frame distance is unreliable.

Let $e$ be the map endpoint and integer $u\ge0$ the index span in half-frame
units. For $L>1$, affine progression $j_i\simeq b+\rho i$ satisfies
$b=e-\rho(L-1)$. Eliminating $b$ and quantizing gives
\begin{equation}
\begin{aligned}
\rho&=\frac{u}{2(L-1)},\\
j_i(e,u)&=e-\operatorname{round}\!\left(\rho(L-1-i)\right).
\end{aligned}
\label{eq:alignment}
\end{equation}
Bounds $0\le e<N'$ and $0\le u\le2e$, with nonnegative half-integers rounded
up, ensure $0\le j_0\le\cdots\le j_{L-1}<N'$. Rate $\rho$ reflects both
motion and keyframe sampling, not metric velocity. Crucially,
\begin{equation}
j_{L-1}(e,u)=e \quad \text{for every admissible }u.
\label{eq:endpoint}
\end{equation}
Rates below/above one repeat/skip map indices; $u=0$ repeats one map frame.
For $L=5$, $e=8$, rates $\rho=0.5$ and $\rho=2$ yield $(6,6,7,7,8)$ and
$(0,2,4,6,8)$. Both end at the current query's assigned frame, although
that assignment may still be incorrect.

For each endpoint in $\mathcal C_{L-1}$, we propose stationary and feasible
unit-rate models. Candidates $c\in\mathcal C_i$ in the first half-window,
$0\le i<\lceil L/2\rceil$, $i<L-1$, supply further spans by solving
$e-c\simeq\rho(L-1-i)$:
\begin{equation}
u=\operatorname{round}\!\left(
\frac{2(e-c)(L-1)}{L-1-i}\right),\qquad
c\le e,\quad 0\le u\le2e.
\label{eq:rateproposal}
\end{equation}
Duplicate models are removed. Early queries provide a longer baseline
and avoid small temporal denominators. Rates have resolution $1/[2(L-1)]$,
but only candidate-derived and default models are searched, without a
prespecified speed interval or exhaustive grid. Each window assumes
approximately constant forward advancement, re-estimated at each invocation.

\subsection{Index-Tolerant Voting and Hypothesis Selection}
For a candidate model, measure the distance from its predicted map
index to the nearest retrieved candidate, clipped at $R+1$:
\begin{equation}
\delta_i(e,u)=\min\!\left(R+1,
\min_{c\in\mathcal C_i}|c-j_i(e,u)|\right).
\label{eq:residual}
\end{equation}
An empty candidate set gives $\delta_i=R+1$. The vote count is
\begin{equation}
V(e,u)=\sum_{i=0}^{L-1}\ind[\delta_i(e,u)\le R].
\label{eq:voting}
\end{equation}
Each query contributes at most one vote regardless of candidate clustering;
$R$ accommodates local deviations and index quantization.

Hypotheses require $V(e,u)\ge\lceil L/2\rceil$ and are ranked by decreasing
votes, then increasing $\sum_i\delta_i$ and $d(Q_{L-1},P_e)$; index ordering
breaks remaining ties. Proposals within $R$ endpoint indices and $2R$ span
units are suppressed, leaving up to $M$ models in coarse set $\mathcal H$.
If the correct endpoint is absent from the newest shortlist, local
refinement may not recover it.

\subsection{Aligned Fine Verification and Current-Pose Recovery}
Fine verification checks descriptor support for the ordered pairs proposed
by voting. For each $(e,u)\in\mathcal H$, it searches integer models
$(e',u')\in\mathbb Z^2$ in
\begin{equation}
\begin{split}
\mathcal F(e,u)=\{(e',u'):\;&|e'-e|\le R,\ |u'-u|\le2R,\\
&0\le e'<N',\ 0\le u'\le2e'\}.
\end{split}
\label{eq:refine}
\end{equation}
Shared models are evaluated once. Each aligned pair has distance
$d_i=d(Q_i,P_{j_i(e',u')})$ and best column shift
$a_i=a(Q_i,P_{j_i(e',u')})$. The newest distance must be finite and satisfy
$d_{L-1}\le\tau_{\rm seq}$. Contributing pairs require finite distances
and shifts consistent with this anchor:
\begin{equation}
\begin{aligned}
\Delta_i&=|a_i-a_{L-1}|,\\
\mathcal V(e',u')&=\{i:d_i\text{ is finite},\ \Delta_i\le2\}.
\end{aligned}
\label{eq:shiftgate}
\end{equation}
This two-column shift-consistency test rejects incompatible descriptor alignments
even when indices agree; it is not an absolute-heading measurement. Define
\begin{equation}
\begin{aligned}
D(e',u')&=\frac{1}{|\mathcal V(e',u')|}
\sum_{i\in\mathcal V(e',u')}d_i,\\
&\hspace{12pt}|\mathcal V(e',u')|\ge\lceil L/2\rceil.
\end{aligned}
\label{eq:seqscore}
\end{equation}
Normalization uses contributing queries, not map span, so repeated map
frames remain valid at $u=0$ without duplicating a query's contribution.
Apart from the newest pair, individual distances need not pass
$\tau_{\rm seq}$. Acceptance requires the latest-frame gate, minimum support
and $D(e',u')\le\tau_{\rm seq}$. We minimize $D$, then latest-frame distance,
yielding $(e^\star,u^\star)$ with score $D^\star$.

The endpoint identity in \eqref{eq:endpoint} determines the registration
pair: the current scan $Q_{L-1}$ is aligned to $P_{e^\star}$, giving
\begin{equation}
\widehat T_{\mathrm{map}\leftarrow Q_{L-1}}
=T_{e^\star}T^{\mathrm{ICP}}_{P_{e^\star}\leftarrow Q_{L-1}}.
\label{eq:pose}
\end{equation}
ICP maps the latest static cloud to the keyframe; $T_{e^\star}$ places it
in the map. It starts from identity, without descriptor-shift initialization.
After convergence, the pose is published with the current timestamp,
original keyframe ID and $D^\star$. Convergence does not certify the association.

\par\smallskip
\noindent\begin{minipage}{\columnwidth}
\hrule\vspace{4pt}
\small
\refstepcounter{algorithm}\label{alg:seq_dynarc}
\textbf{Algorithm~\thealgorithm: Seq\_DynaRC inference}\\[2pt]
\textit{Input:} query window $\mathcal W_t$, map $\mathcal M$,
$K,M,R,\tau_{\rm seq}$.\\
\textit{Output:} current pose and sequence score, or no result.
\begin{enumerate}
\setlength{\itemsep}{1pt}\setlength{\parskip}{0pt}
\item Retrieve and rerank candidates to obtain each $\mathcal C_i$.
\item For $e\in\mathcal C_{L-1}$, propose spans by \eqref{eq:rateproposal},
plus stationary and feasible unit-rate models.
\item Vote by \eqref{eq:voting}; require half-window support, rank and
suppress nearby models to form $\mathcal H$.
\item Evaluate distinct models in $\bigcup_{h\in\mathcal H}\mathcal F(h)$.
Keep those passing latest-distance, shift-support and mean-distance tests
\eqref{eq:shiftgate}--\eqref{eq:seqscore}.
\item If none survives, return no result. Otherwise minimize $(D,d_{L-1})$
lexicographically and register $Q_{L-1}$ to $P_{e^\star}$.
Return \eqref{eq:pose} and $D^\star$ only if ICP converges.
\end{enumerate}
\vspace{2pt}\hrule
\end{minipage}\par\smallskip

For $H_0$ proposed models, voting costs $O(H_0LK)$. Fine verification
visits at most $M(2R+1)(4R+1)L$ descriptor pairs before boundary pruning
and caching shared comparisons. Each comparison costs
$O(|\mathcal L|N_rN_s)$ after sector-key alignment. Measured runtime must
also include tree maintenance, retrieval, shift initialization and ICP.

\section{Experiments}
\label{sec:experiments}
\subsection{Sensor Platform and Datasets}
\begin{figure*}[!t]
\centering
\includegraphics[width=\textwidth]{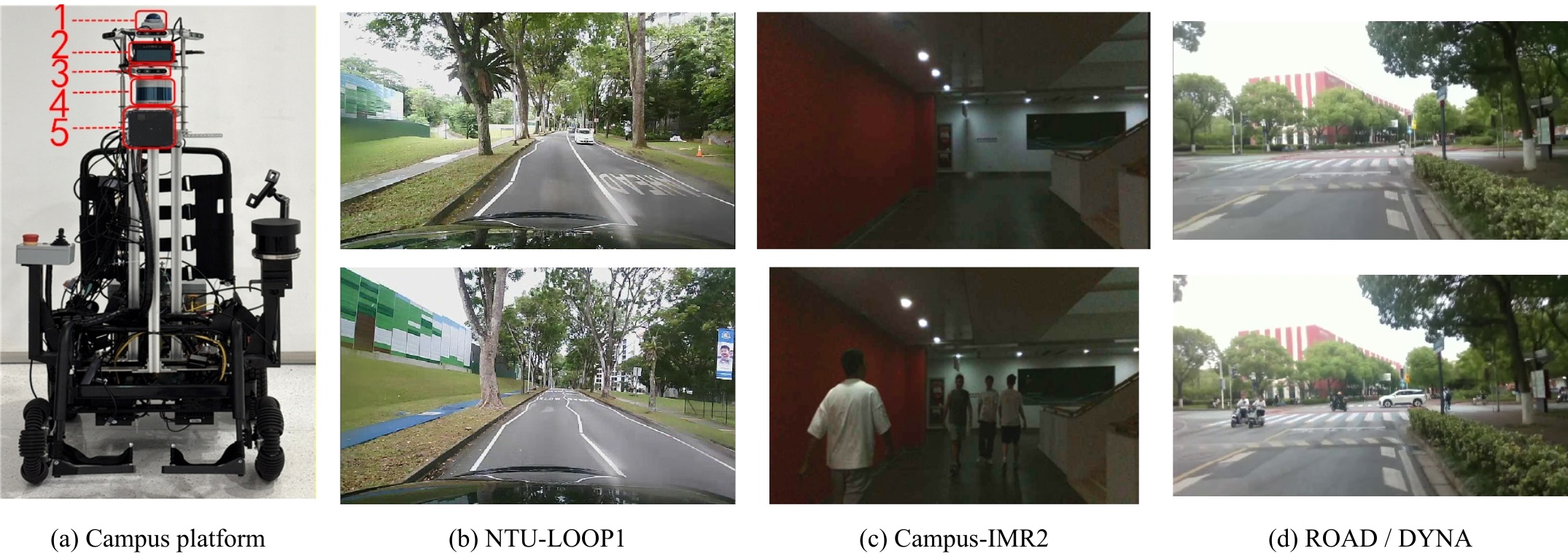}
\caption{Platform and scenes. (a) Campus platform: 1, Livox MID-360;
2, Livox HAP; 3, RealSense D435i; 4, Velodyne HDL-32; 5, ZF FRGen21.
(b) NTU-LOOP1. (c) IMR2 with/without pedestrians. (d) ROAD/DYNA at
15:30/18:00 (top/bottom). Platform and corridor photographs are from
the RaVI-SLAM experiments~\cite{RaVISLAM}.}
\label{fig:acquisition}
\end{figure*}

We use NTU4DRadLM~\cite{NTU} and our Campus recordings
(Fig.~\ref{fig:acquisition}). NTU uses an Oculii EAGLE 4D radar:
LOOP1 is a road traversal, and CP/SMOKE pairs clear and smoke-affected
sessions. The RaVI-SLAM wheelchair platform~\cite{RaVISLAM} carries a
ZF FRGen21 radar, RealSense D435i camera/IMU, and LiDAR sensors.
Campus-IMR2 contains repetitive corridors and pedestrian occlusions;
ROAD/DYNA revisits a campus road under different traffic conditions.
Relocalization consumes radar only; other sensors support offline mapping,
reference construction and alignment, while images illustrate the scenes.

\begin{figure*}[!t]
\centering
\includegraphics[width=\textwidth]{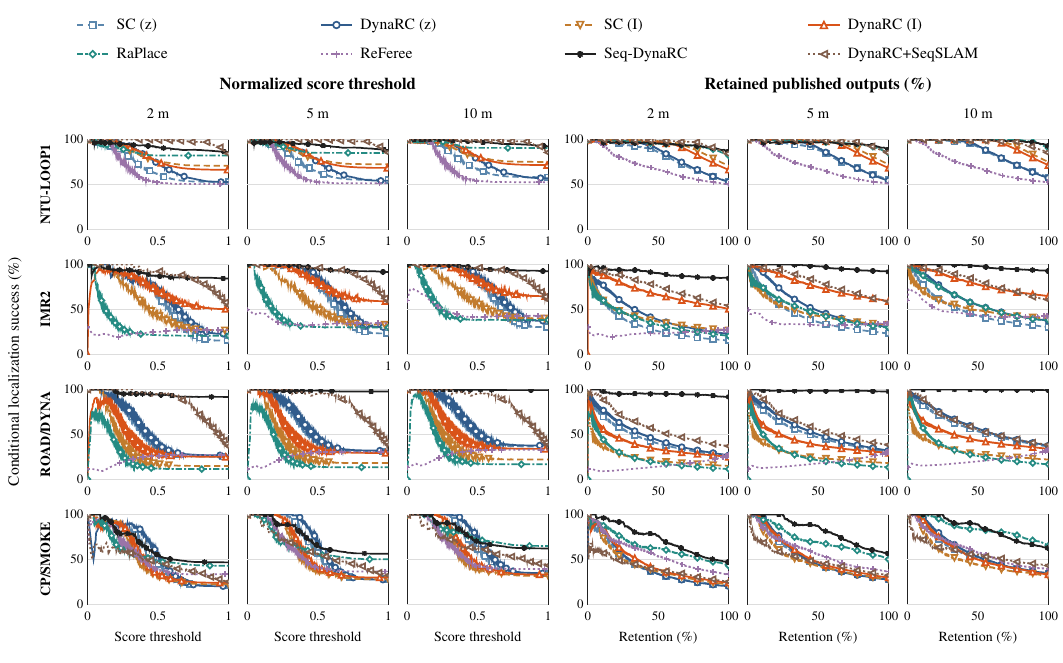}
\caption{Conditional XY localization success within 2, 5 and 10\,m.
Left: per-run min--max normalized score threshold. Right: fraction of
published outputs retained by the score gate, including 100\% retention
of each method's own outputs. Both sequence methods use intensity DynaRC
descriptors. $z$ and I denote height and intensity.}
\label{fig:comparison}
\end{figure*}

\subsection{Evaluation Protocol}
All methods search the same map keyframes. For LOOP1 and IMR2,
RaVI-SLAM reconstructs both loops in a common frame: loop one supplies
the map and loop two the queries. For ROAD/DYNA and CP/SMOKE,
startup LiDAR registration and calibrated LiDAR--radar extrinsics align
the separate sessions. Reference positions and orientations are interpolated
at output timestamps; out-of-range times and gaps exceeding 1\,s are excluded.
These references inherit reconstruction, calibration and alignment error.

We compare SC and DynaRC with height/intensity channels, Seq\_DynaRC,
and point-cloud adaptations of RaPlace~\cite{RaPlace} and ReFeree~\cite{ReFeree}.
For RaPlace, points populate a Cartesian bird's-eye-view image using
maximum intensity/RCS per cell, followed by area-based downsampling.
For ReFeree, returns populate a binary range--azimuth grid for polar
descriptor extraction; repeated returns in one cell count once.
Both adapters retrieve keyframes and use point-to-point ICP for pose
recovery. These results evaluate the 4D point-cloud adaptations.
The added DynaRC+SeqSLAM baseline uses the same intensity descriptors,
masked distances, searchable map and final ICP as Seq\_DynaRC. It applies
local contrast normalization to the full query--map distance matrix,
searches forward constant-rate paths, and scores the best path relative
to the best endpoint outside a map-index exclusion neighborhood. Its newest query
also anchors the output endpoint. Thus, the comparison concerns candidate
search, rate proposals, scoring and rejection, rather than the ability to
vary rate alone. Table~\ref{tab:eval_params} specifies the settings.

\begin{table}[t]
\centering
\caption{Default evaluation settings (map-index units unless stated).}
\label{tab:eval_params}
\footnotesize
\setlength{\tabcolsep}{3pt}
\begin{tabular}{@{}>{\raggedright\arraybackslash}p{0.30\columnwidth}>{\raggedright\arraybackslash}p{0.66\columnwidth}@{}}
\toprule
Setting & Value\\
\midrule
Polar grid / channels & $20\times20$; $120^\circ$; SC/DynaRC: $z$ or I; both sequence methods: I\\
Maximum radius & LOOP1: 250\,m; IMR2: 50\,m; ROAD/DYNA: 250\,m; CP/SMOKE: 150\,m\\
Mask / norm tolerance & $n_{\min}=5$; $\lambda=4/9$; $\epsilon=10^{-6}$\\
Sequence schedule & Window $L=30$; stride 5 processed scans\\
Seq\_DynaRC search & 25 candidates reranked to $K=5$; $M=5$ models; tolerance $R=3$\\
Verification & Support $\ge15$; shift tolerance 2 columns; $\tau_{\rm seq}=1$\\
SeqSLAM search & Rate $0.2{:}0.05{:}5.0$ map frames/query; contrast/exclusion radii: 50/10; ratio gate: 1\\
Final ICP & Correspondence limit 2\,m; 50 iterations; transformation/fitness tolerances $10^{-6}$\\
Single-frame ablation & I channel; median key; relaxed mask: $n_{\min}=2$, $\lambda=3/8$, including dynamic-only cells\\
\bottomrule
\end{tabular}
\end{table}

We distinguish publication availability $N_{\rm pub}/N_{\rm att}$ from
score retention among published outputs. Figure~\ref{fig:comparison} sweeps
the native gate $s<\tau$; min--max scaling is for display and does not
calibrate confidence across methods. Success is the fraction of retained,
reference-valid poses whose XY error is below 2, 5 or 10\,m. Retention uses
all recorded outputs, including those without valid references. Single-frame
methods run per processed scan; sequence methods use the schedule in
Table~\ref{tab:eval_params}.
Actual sampling differs between runs, so the curves describe system output
quality, not performance on identical query histories.

\subsection{Comparison across Environments}
At full score retention, Seq\_DynaRC has the highest conditional 2\,m success
on all pairs: 86.0\%, 84.4\%, 91.8\% and 46.6\% on LOOP1, IMR2,
ROAD/DYNA and CP/SMOKE; DynaRC+SeqSLAM obtains 83.8\%, 54.3\%, 36.8\%
and 25.2\%. On ROAD/DYNA, Seq\_DynaRC publishes 857/3,127 attempts
(27.4\%). The other 2,270 attempts end before ICP: the temporal matcher
has no accepted hypothesis even with the distance gate set to one.
Limited overlap, dynamic occlusion and repeated structures can remove
consistent candidate support; the logs establish rejection counts, not
the cause of every rejected window. This gate trades availability for
reliability, whereas SeqSLAM publishes all 1,991 of its processed attempts.

An exact-timestamp comparison using the existing CSVs
(Table~\ref{tab:shared}) retains the Seq\_DynaRC advantage on three pairs;
the two methods are close on LOOP1. This reduces output-time differences,
but conditions on joint publication and does not equalize preceding windows.
The results support the complete search-and-rejection design; they do not
isolate candidate pruning from every other difference in sequence matching.

\begin{table}[t]
\centering
\caption{2\,m success (\%) on exact shared output timestamps, without score
filtering. $N_\cap$ counts reference-valid poses published by both methods.}
\label{tab:shared}
\footnotesize
\setlength{\tabcolsep}{5pt}
\begin{tabular}{@{}lrrr@{}}
\toprule
Pair & $N_\cap$ & Seq\_DynaRC & DynaRC+SeqSLAM\\
\midrule
NTU-LOOP1 & 76 & 86.84 & 88.16\\
Campus-IMR2 & 145 & 88.28 & 66.21\\
ROAD/DYNA & 112 & 91.96 & 50.89\\
CP/SMOKE & 383 & 55.09 & 40.21\\
\bottomrule
\end{tabular}
\end{table}

\begin{table}[t]
\centering
\caption{IMR2 ablations with intensity descriptors and median keys.
XY success (\%) at full score retention; $N$: reference-valid outputs.
In (b), the full system is the run in Fig.~\ref{fig:comparison};
the other sequence variants disable vote filtering.}
\label{tab:ablation}
\footnotesize
\setlength{\tabcolsep}{3pt}
\begin{tabular}{@{}lrrrr@{}}
\toprule
Variant & $N$ & 2\,m & 5\,m & 10\,m\\
\midrule
\multicolumn{5}{@{}l}{\textbf{(a) Single-frame dynamic handling}}\\[2pt]
DynaRC & 5742 & 48.90 & 58.06 & 64.09\\
w/o joint module & 5742 & 34.81 & 42.95 & 50.68\\
\midrule
\multicolumn{5}{@{}l}{\textbf{(b) Sequence matching}}\\[2pt]
Full system & 778 & 84.45 & 91.77 & 92.67\\
w/o vote filter & 1078 & 65.40 & 73.19 & 76.81\\
w/o vote filter, rate & 1080 & 54.44 & 64.17 & 68.52\\
w/o vote filter, fine & 1073 & 64.96 & 73.53 & 77.07\\
w/o vote filter, rate, fine & 1076 & 56.32 & 63.38 & 67.57\\
\bottomrule
\end{tabular}
\end{table}

Channel choice is scene dependent. Intensity DynaRC exceeds height DynaRC
on IMR2 at 2\,m (50.4\% versus 22.7\%) and LOOP1 (66.2\% versus 52.6\%),
but slightly trails it on ROAD/DYNA (24.8\% versus 26.7\%). The benefit
therefore cannot be attributed to one universally preferable channel.
On CP/SMOKE, Seq\_DynaRC exceeds RaPlace at 5\,m (56.2\% versus 50.0\%),
whereas RaPlace leads at 10\,m (65.0\% versus 62.2\%).

\subsection{Component Ablations}
\label{sec:ablations}
Table~\ref{tab:ablation}(b) uses the full-system run from
Fig.~\ref{fig:comparison}: 785/1,082 attempts publish poses, with 778 valid
references. Removing the vote filter retains hypothesis voting and fine
verification but forces output after warm-up: 1,085/1,085 poses, 1,078
evaluable. Full-system $N$ is smaller because of online rejection, not
offline filtering. Its 2\,m success is 84.45\% versus 65.40\% without the
filter, on different scheduled timestamps.

Further variants disable rate adaptation ($\rho=1$), fine verification,
or both. Without vote filtering, fixing the rate costs 10.95 percentage
points at 2\,m; removing fine verification changes success by less than one
point at each tolerance. This tests unequal index advancement, not controlled
acceleration.

In (a), we jointly restore dynamic descriptor inputs and disable masking,
retaining median keys, denoising and static-cloud ICP. The full variant uses
the relaxed mask in Table~\ref{tab:eval_params}. Both publish all 5,780 common
queries (5,742 reference-valid). Joint handling gains 14.09/15.12/13.41
percentage points at 2/5/10\,m. Among 5,248 valid queries with dynamic-point
fraction $\ge0.2$, 2\,m success increases from 32.24\% to 46.70\%.

\emph{Final registration:} on identical published Seq\_DynaRC samples, 2\,m
success before/after ICP is 76.6\%/86.0\% on LOOP1, 83.2\%/84.4\% on IMR2,
76.6\%/91.8\% on ROAD/DYNA and 29.1\%/46.6\% on CP/SMOKE; ``before'' uses
the retrieved keyframe position. ICP substantially refines poses, but these
statistics measure neither all-query convergence nor yaw accuracy. Our
contribution is place association, followed by standard ICP.

\subsection{Computational Efficiency}
Table~\ref{tab:runtime} reports wall-clock time from descriptor construction
to rejection or publication on an Intel Core i7-1360P with one ROS callback thread.
Only the selected endpoint undergoes ICP. Input rates use processed-scan
timestamps; costs are per attempt. LOOP1 timing and accuracy runs are separate.

\begin{table*}[t]
\centering
\caption{Runtime in ms. Input Hz is the processed-scan rate in dataset time;
Out/att. is publication count/attempt count. Descriptor, retrieval and
sequence means include rejected attempts; ICP/call includes executed calls
only. Total includes bookkeeping; P95 is the 95th percentile.}
\label{tab:runtime}
\footnotesize
\setlength{\tabcolsep}{4pt}
\begin{tabular}{@{}llrrrrrrrr@{}}
\toprule
Pair & Method & Out/att. & Input Hz & Descr. & Retrieval & Sequence & ICP/call & Total mean & Total P95\\
\midrule
NTU-LOOP1 & SC-$z$ & 3115/3115 & 10.83 & 0.11 & 0.15 & --- & 65.22 & 65.52 & 148.34\\
 & DynaRC (I) & 3197/3197 & 11.12 & 0.20 & 0.19 & --- & 59.00 & 59.42 & 138.25\\
 & Seq\_DynaRC & 547/673 & 11.79 & 0.24 & 10.29 & 4.18 & 77.06 & 77.50 & 193.52\\
\midrule
Campus-IMR2 & SC-$z$ & 5458/5458 & 15.74 & 0.02 & 0.10 & --- & 1.17 & 1.31 & 2.53\\
 & DynaRC (I) & 5449/5449 & 15.70 & 0.03 & 0.15 & --- & 0.68 & 0.86 & 1.66\\
 & Seq\_DynaRC & 785/1082 & 15.67 & 0.03 & 12.58 & 4.34 & 0.76 & 17.54 & 25.18\\
\midrule
ROAD/DYNA & SC-$z$ & 14815/14815 & 15.76 & 0.03 & 0.15 & --- & 4.38 & 4.75 & 10.53\\
 & DynaRC (I) & 15662/15662 & 16.67 & 0.05 & 0.18 & --- & 3.53 & 3.84 & 8.28\\
 & Seq\_DynaRC & 857/3127 & 16.67 & 0.05 & 17.60 & 1.65 & 2.52 & 20.06 & 29.80\\
 & DynaRC+SeqSLAM & 1991/1991 & 10.62 & 0.05 & 252.04 & 6.00 & 2.92 & 261.06 & 297.02\\
\midrule
CP/SMOKE & SC-$z$ & 4591/4591 & 11.40 & 0.11 & 0.06 & --- & 59.08 & 59.27 & 121.56\\
 & DynaRC (I) & 4585/4585 & 11.38 & 0.19 & 0.09 & --- & 58.51 & 58.82 & 120.88\\
 & Seq\_DynaRC & 952/958 & 11.96 & 0.26 & 9.62 & 5.74 & 68.55 & 83.80 & 148.77\\
\bottomrule
\end{tabular}
\end{table*}

NTU's Oculii scans are denser than Campus ZF scans~\cite{TransLoc4D}:
Seq\_DynaRC ICP costs 69--77\,ms per call on NTU and 0.76--2.52\,ms on Campus,
where retrieval dominates. Overlap and convergence also affect cost.
On ROAD/DYNA, full-matrix SeqSLAM retrieval takes 252.04\,ms per attempt
versus 17.60\,ms for Seq\_DynaRC; totals are 261.06 and 20.06\,ms. The SeqSLAM
trace matches all 1,991 published timestamps. Its lower processed input
rate (10.62 versus 16.67\,Hz) gives 30-frame spans of about 2.7 versus 1.7\,s.
Thus, equal stride does not equalize observations. These timings establish
lower search cost, while both sampling and rejection affect accuracy.

\section{Discussion and Limitations}
Under the static-map assumption, occluding distinctive returns can worsen
query agreement. Masked sectors are excluded, however, so distance need not
increase monotonically. No usable sector yields distance one: this penalizes
missing overlap but remains finite, allowing the pair to contribute to
shift-consistent support and the sequence mean. Heavy occlusion and scene
changes therefore remain ambiguous.

\section{Conclusion}
Seq\_DynaRC uses motion labels to compare static scene evidence and
candidate-supported sequence votes to associate the current scan.
The evaluation jointly characterizes localization quality, output
availability and search cost. Adaptive advancement improves forced-output
endpoint selection,
while joint static filtering and masking improve single-frame
success with the median fixed. Final ICP remains responsible for local
pose refinement; channel choice and lost overlap limit generalization.

\IEEEtriggeratref{22}
\bibliographystyle{IEEEtran}
\bibliography{references}
\end{document}